\documentclass[10pt,twocolumn,letterpaper]{article}

\usepackage{cvpr}
\usepackage{times}
\usepackage{epsfig}
\usepackage{graphicx}
\usepackage{amsmath}
\usepackage{amssymb}
\usepackage[]{algorithm2e}
\usepackage{color,soul} % for \hl

\usepackage[pagebackref=true,breaklinks=true,letterpaper=true,colorlinks,bookmarks=false]{hyperref}

\usepackage{url}
\usepackage{graphicx}
\graphicspath{{imgs/}}
\DeclareGraphicsExtensions{.pdf,.jpg,.png}
\usepackage{xcolor}

\usepackage{enumitem}
\setlist{nolistsep}

\newcommand{\fakesubsection}[1]{\smallskip\noindent\textbf{#1:}}

\cvprfinalcopy % *** Uncomment this line for the final submission

\def\cvprPaperID{60} % *** Enter the CVPR Paper ID here

\ifcvprfinal\fi
\begin{document}

%%%%%%%%% TITLE
%\title{A Unified View of Action Segmentation and Detection}
\title{Temporal Convolutional Networks\\ for Action Segmentation and Detection}
%\title{Action Segmentation or Detection?\\ The Difference is in the Metrics}

%\author{Colin Lea \and Austin Reiter \and Ren\'{e} Vidal \and Gregory D. Hager \\
%Johns Hopkins University
\author{Colin Lea 
	\hspace{11px} Michael D. Flynn 
	\hspace{11px} Ren\'{e} Vidal  
	\hspace{11px} Austin Reiter
	\hspace{11px} Gregory D. Hager  \\	
Johns Hopkins University \\
	 \{clea1@, mflynn@, rvidal@cis., areiter@cs., hager@cs.\}jhu.edu
%3400 N. Charles St, Baltimore, MD, 21218\\
%{\tt\small clea1@jhu.edu}
%\and Austin Reiter 
%\and Ren\'{e} Vidal
%\and Gregory D. Hager
% For a paper whose authors are all at the same institution,
% omit the following lines up until the closing ``}''.
% Additional authors and addresses can be added with ``\and'',
% just like the second author.
}

%\author{Colin lea\\
%Institution1\\
%Institution1 address\\
%{\tt\small clea1@jhu.edu}
%% For a paper whose authors are all at the same institution,
%% omit the following lines up until the closing ``}''.
%% Additional authors and addresses can be added with ``\and'',
%% just like the second author.
%% To save space, use either the email address or home page, not both
%\and
%Austin Reiter\\
%Institution2\\
%First line of institution2 address\\
%{\tt\small secondauthor@i2.org}
%\and
%Rene Vidal\\
%Institution2\\
%First line of institution2 address\\
%{\tt\small secondauthor@i2.org}
%\and
%Gregory D. Hager\\
%Institution2\\
%First line of institution2 address\\
%{\tt\small secondauthor@i2.org}
%}

\maketitle
%\thispagestyle{empty}

%%%%%%%%% ABSTRACT
% As a general rule, do not put math, special symbols or citations in the abstract or keywords.
\begin{abstract}

The ability to identify and temporally segment fine-grained human actions throughout a video is crucial for robotics, surveillance, education, and beyond.
Typical approaches decouple this problem by first extracting local spatiotemporal features from video frames and then feeding them into a temporal classifier that captures high-level temporal patterns.
We introduce a new class of temporal models, which we call Temporal Convolutional Networks (TCNs), that use a hierarchy of temporal convolutions to perform fine-grained action segmentation or detection.
Our Encoder-Decoder TCN uses pooling and upsampling to efficiently capture long-range temporal patterns whereas our Dilated TCN uses dilated convolutions. 
We show that TCNs are capable of capturing action compositions, segment durations, and long-range dependencies, and are over a magnitude faster to train than competing LSTM-based Recurrent Neural Networks.
We apply these models to three challenging fine-grained datasets and show large improvements over the state of the art.

\end{abstract}
\section{Introduction}
\label{sec:intro}

\begin{figure}
	\centering
	\includegraphics[width=\hsize]{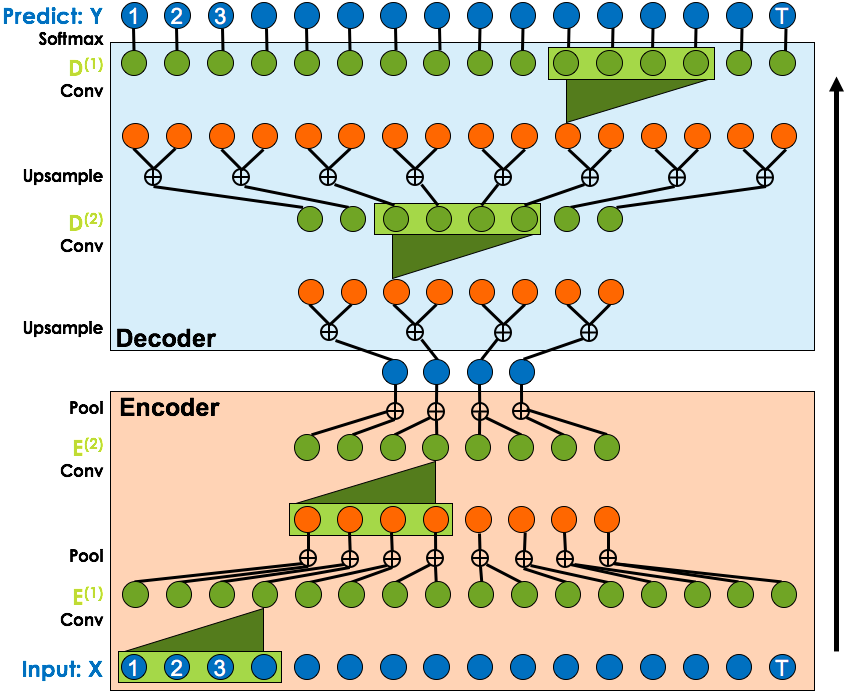}
	\caption{Our Encoder-Decoder Temporal Convolutional Network (ED-TCN) hierarchically models actions using temporal convolutions, pooling, and upsampling.}
	\label{fig:ED_TCN}
\end{figure}

Action segmentation is crucial for applications ranging from collaborative robotics to analysis of activities of daily living. Given a video, the goal is to simultaneously segment every action in time and classify each constituent segment. 
In the literature, this task goes by either action segmentation or detection. 
We focus on modeling situated activities -- such as in a kitchen or surveillance setup -- which are composed of dozens of actions over a period of many minutes. 
These actions, such as cutting a tomato versus peeling a cucumber, are often only subtly different from one another.

% within a structured task such as cooking. 
% \textit{Situated} implies that all videos in a given dataset only come from a single domain (e.g., cooking) and \textit{fine-grained} implies that each of the action classes may be subtly different from one another (e.g., cutting tomato versus peeling cucumber).

Current approaches decouple this task into extracting low-level spatiotemporal features and applying high-level temporal classifiers. 
While there has been extensive work on the former, recent temporal models have been limited to sliding window action detectors~\cite{rohrbach_ijcv_2015,singh_cvpr_2016_merl,ni_cvpr_2016}, which typically do not capture long-range temporal patterns; segmental models~\cite{richard_cvpr_2016,lea_eccv_2016,pirsiavash_cvpr_2014}, which capture within-segment properties but assume conditional independence, thereby ignoring long-range latent dependencies;
and recurrent models~\cite{singh_cvpr_2016_merl,huang_eccv_2016}, which can capture latent temporal patterns 
but are difficult to interpret, have been found empirically to have a limited span of attention~\cite{singh_cvpr_2016_merl}, and are hard to correctly train~\cite{pascanu_icml_2013}. 

In this paper, we discuss a class of time-series models, which we call Temporal Convolutional Networks (TCNs), 
% that are able to capture complex temporal patterns not easily modeled by CRFs or RNNs. 
that overcome the previous shortcomings by capturing long-range patterns using a hierarchy of temporal convolutional filters. 
% These models are characterized by their use of temporal convolutions, layer-wise computations, and large receptive fields.
We present two types of TCNs:
First, our Encoder-Decoder TCN (ED-TCN)
% \footnote{Reviewer note: Our model first appeared in a short workshop paper. CVPR guidelines state this can be re-used for a conference submission.} 
only uses a hierarchy of temporal convolutions, pooling, and upsampling but can efficiently capture long-range temporal patterns.
The ED-TCN has a relatively small number of layers (e.g., 3 in the encoder) but each layer contains a set of long convolutional filters.
% capture low-, intermediate-, and high-level temporal patterns. 
% Convolution filters in each layer capture how features at lower levels change over time and pooling enables efficient computation of long-range temporal patterns.
%, and normalization improves robustness towards various environmental conditions.
Second, a Dilated TCN uses dilated convolutions instead of pooling and upsampling and adds skip connections between layers.
% to capture long-range patterns.
% \footnote{Dilated convolutions are also sometimes called atrous convolutions.}
% with exponentially increasing dilation rates to achieve a large receptive field without using any pooling operations.
This is an adaptation of the recent WaveNet~\cite{wavenet} model, which shares similarities to our ED-TCN but was developed for speech processing tasks.
The Dilated TCN has more layers, but each uses dilated filters that only operate on a small number of time steps.
% which effectively operate over a long period of time a large effective receptive field but
Empirically, we find both TCNs are capable of capturing features of segmental models, such as action durations and pairwise transitions between segments, as well as long-range temporal patterns similar to recurrent models. 
These models tend to outperform our Bidirectional LSTM (Bi-LSTM)~\cite{bi_lstm} baseline and are over a magnitude faster to train. 
The ED-TCN in particular produces many fewer over-segmentation errors than other models. 

In the literature, our task goes by the name action segmentation \cite{fathi_cvpr_2013,fathi_cvpr_2011,fathi_iccv_2011,kuehne_wacv_2016,singh_cvpr_2016_ego,lea_eccv_2016,huang_eccv_2016} or action detection~\cite{singh_cvpr_2016_merl,ni_cvpr_2014,ni_cvpr_2016,richard_cvpr_2016}. 
% Unlike in areas such object recognition, where there is a clear distinction between segmentation and detection, for fine-grained actions they are the same. 
% papers treat them as separate problems 
% However, they are treated as separate problems
Despite effectively being the same problem, the temporal methods in segmentation papers tends to differ from detection papers, as do the metrics by which they are evaluated. 
% and will self-identify as approaching one or the other, 
% despite the fact that it is trivial to go back and forth between a set of detections and segmentations. 
% The temporal of methods in these papers tend to differ, as do the metrics in which they are evaluated. 
In this paper, we evaluate on datasets targeted at both tasks and propose a segmental F1 score, which we qualitatively find is more applicable to real-world concerns for both segmentation and detection than common metrics. 
We use MERL Shopping which was designed for action detection, Georgia Tech Egocentric Activities which was designed for action segmentation, and 50 Salads which has been used for both.

Code for our models and metrics, as well as dataset features and predictions,\footnote{The release of the MERL Shopping features depends on the permission of the original authors. All other features will be available.} will be released upon acceptance.

%we take a very different viewpoint than much of the time-series and video communities 
%which is largely directed towards RNN-type models,
%and instead frame the problem of action segmentation in a similar lens as semantic segmentation where the goal is to densely label every pixel in an image with a label. 

\section{Related Work}
\label{sec:related}

Action segmentation methods predict what action is occurring at every frame in a video and detection methods output a sparse set action segments, where a segment is defined by a start time, end time, and class label. It is possible to convert between a given segmentation and set of detections by simply adding or removing null/background segments.

\fakesubsection{Action Detection}
Many fine-grained detection papers use sliding window-based detection methods on spatial or spatiotemporal features. 
Rohrbach \etal \cite{rohrbach_ijcv_2015} used Dense Trajectories~\cite{wang_iccv_2013} and human pose features on the MPII Cooking dataset. At each frame they evaluated a sliding SVM for many candidate segment lengths and performed non-maximal suppression to find a small set of action predictions.
% At each frame, they computed action scores for a large number of candidate action durations and performed non-maximal suppression to find a small set of action predictions.
% The scores for each detection interval are a sum of the per-frame scores.
Ni \etal~\cite{ni_cvpr_2016} used an object-centric feature representation, which iteratively parses object locations and spatial configurations, and applied it to the MPII Cooking and ICPR 2012 Kitchen datasets. Their approach used Dense Trajectory features as input into a sliding-window detection method with segment intervals of 30, 60, and 90 frames.
Singh \etal \cite{singh_cvpr_2016_merl} improved upon this by feeding per-frame CNN features into an LSTM model and applying a method analogous to non-maximal suppression to the LSTM output.
We use Singh's proposed dataset, MERL Shopping, and show our approach outperforms their LSTM-based detection model.
% to compute a sparse set of action segments. 
% Singh \etal evaluated on MERL Shopping, their proposed action detection dataset, and MPII Cooking.
Recently, Richard \etal \cite{richard_cvpr_2016} introduced a segmental approach that incorporates a language model, which captures pairwise transitions between segments, and a duration model, which ensures that segments are of an appropriate length. In the experiments we show that our model is capable of capturing both of these components. 
% sequential actions maximizes over many candidate action durations by . 
% This was evaluated on the THUMOS2014 and 50 Salads datasets.

Some of these datasets, including MPII Cooking, have been used for classification (e.g., \cite{cheron_iccv_2015,zhou_cvpr_2015}), however, this task assumes the boundaries of each segment are known.

\fakesubsection{Action Segmentation}
Segmentation papers tend to use temporal models that capture high-level patterns, for example RNNs or Conditional Random Fields (CRFs).
% , as opposed to fixed-interval detection models.
The line of work by Fathi \etal~\cite{fathi_cvpr_2011,fathi_cvpr_2013,fathi_iccv_2011} used a segmental model that captured object states at the start and end of each action (e.g., the appearance of bread before and after spreading jam). They applied their work to the Georgia Tech Egocentric Activities (GTEA) dataset, which we use in our experiments.
Singh \etal ~\cite{singh_cvpr_2016_ego} used an ensemble of CNNs to extract egocentric-specific features on the GTEA dataset but did not use a high-level temporal model.
Lea \etal \cite{lea_eccv_2016} introduced a spatiotemporal CNN with a constrained segmental model which they applied to 50 Salads. Their model reduced the number of action over-segmentation errors by constraining the maximum number of candidate segments. We show our TCNs produce even fewer over-segmentation errors.
Kuehne \etal \cite{kuehne_cvpr_2014,kuehne_wacv_2016} modeled actions using Hidden Markov Models on Dense Trajectory features, which they applied with a high-level grammar to 50 Salads. 
Other work has looked at semi-supervised methods for action segmentation, such as Huang \etal~\cite{huang_eccv_2016}, which reduces the number of required annotations and improves performance when used with RNN-based models. 
% . They introduced an extension of the CTC loss for RNN-based models which actually improves performance over the fully-supervised case and requires fewer labels. 
It is possible that their approach could be used with TCNs for improved performance. 

\fakesubsection{Large-scale Recognition}
There has been substantial work on spatiotemporal models for large scale video classification and detection~\cite{sun_iccv_2015,jain_cvpr_2015,karpathy_cvpr_2014,simonyan_nips_2014,tran_iccv_2015,peng_thumos_2015,ng_cvpr_2015}. 
These focus on capturing object- and scene-level information from short sequences of images and thus are considered orthogonal to our work, which focuses on capturing longer-range temporal information. The input into our model could be the output of a spatiotemporal CNN. 

\fakesubsection{Other related models}
There are parallels between TCNs and recent work on semantic segmentation, which uses Fully Convolutional CNNs to compute a per-pixel object labeling of a given image. The Encoder-Decoder TCN is most similar to SegNet~\cite{badrinarayanan_arxiv_2015} whereas the Dilated TCN is most similar to the Multi-Scale Context model~\cite{yu_iclr_2016}. TCNs are also related to Time-Delay Neural Networks (TDNNs), which were introduced by Waibel \etal~\cite{Waibel_1990} in the early 1990s. TDNNs apply a hierarchy of temporal convolutions across the input but do not use pooling, skip connections, newer activations (e.g., Rectified Linear Units), or other features of our TCNs.
% introduced time-delay neural networks for speech processing in the early 1990s. These networks are similar to ours in that they apply a hierarchy of temporal convolutions across the input. However, they do not use pooling or upsampling and they applied sigmoid activation functions, which we show do not perform as well as other functions.

% \TODO{Recently X[Povey] \etal~\cite{?} used these for speech recognition. }
% \TODO{Recently Convolutional Autoencoders (CAEs) were used for graphics applications to synthesize human motions from motion capture data ~\ref{holden_siggraph_2016}. 
% It is possible that CAEs could be used to initialize the parameters of our model.}

% , or the activtions that we describe.
% using sigmoid activations between layers. 
% . Interestingly, we find that some of their design decisions, for example the use of sigmoid activations, to provide poor performance as described in our experiments. 
% WaveNet was developed concurrently with our model and is the closest to our work. It was designed for speech processing applications but can be easily adapted for action segmentation as we describe later. 
\section{Temporal Convolutional Networks}
\label{sec:TCN}

In this section we define two TCNs, each of which have the following properties: (1) computations are performed layer-wise, meaning every time-step is updated simultaneously, instead of updating sequentially per-frame (2) convolutions are computed across time, and (3) predictions at each frame are a function of a fixed-length period of time, which is referred to as the receptive field. 
Our ED-TCN uses an encoder-decoder architecture with temporal convolutions and the Dilated TCN, which is adapted from the WaveNet model, uses a deep series of dilated convolutions.

% Computing activations per-layer gives a large speed advantage because all operations can be parallelized on a GPU. 
% Using temporal convolutions enables us to capture how the input features change as a function of time. Note that because our filters are learned, these convolutions can be viewed as 1D or 2D operations. 

The input to a TCN will be a set of video features, such as those output from a spatial or spatiotemporal CNN, for each frame of a given video.
% latent encoding of a spatial CNN applied to each frame or, as we show in the supplemental material, sensor signals (e.g. accelerometers). 
Let $X_t \in \mathbb{R}^{F_0}$ be the input feature vector of length $F_0$ for time step $t$ for $1 \leq t \leq T$. Note that the number of time steps $T$ may vary for each video sequence.
%, and we denote the number of time steps in each layer as $T_l$. 
The action label for each frame is given by vector
% $y_t \in \{1,\dots,C\}$, where $C$ is the number of classes.
$Y_t \in \{0,1\}^C$, where $C$ is the number of classes, such that the true class is $1$ and all others are $0$.

% --------------------------------
\subsection{Encoder-Decoder TCN}
\label{sec:EDTCN}

Our encoder-decoder framework is depicted in Figure~\ref{fig:ED_TCN}.
% , uses temporal convolutions, a non-linear activation function, pooling, and upsampling.
The encoder consists of $L$ layers 
% $E^{(l)} \in \mathbb{R}^{T_{l} \times F_{l}}$,
denotes by $E^{(l)} \in \mathbb{R}^{F_{l} \times T_{l}}$
% for $l \in \{0,\dots,L\}$,
where $F_{l}$ is the number of convolutional filters in a the $l$-th layer and $T_{l}$ is the number of corresponding time steps. 
Each layer consists of temporal convolutions, a non-linear activation function, and max pooling across time. 
% The convolutional filters aim to capture how the input signal changes over time. 
% For clarity, input features $X$ are denoted $E^{(0)}$.

% Each convolution layer is parameterized by a set of $F_{l}$ filters  and corresponding biases. Filters have the same number of channels as their input, $F_{l-1}$, and have filter duration $d$ which is determined via cross validation. We define the $i$-th filter in a layer as matrix $W^{(i)} \in \mathbb{R}^{d \times F_{l-1}}$ with bias vector $b^{(i)}  \in \mathbb{R}$. 
We define the collection of filters in each layer as $W=\{W^{(i)}\}_{i=1}^{F_l}$ for $W^{(i)} \in \mathbb{R}^{d \times F_{l-1}}$ with a corresponding bias vector
% $b^{(i)}  \in \mathbb{R}$ . 
$b  \in \mathbb{R}^{F_l}$. 
% = \{W^{(l,i)}\}_{i=1}^{F_l}
Given the signal from the previous layer, $E^{(l-1)}$, we compute activations $E^{(l)}$ with
% the activations 
% vector for time $t$ is given by
\begin{align}\label{eqn:ED-TCN}
% E^{(l)}_{t,i} = f(\sum_{t'=1}^{d}  W^{(i)}_{t'} E^{(l-1)}_{t-t'+1} + b^{(i)})
E^{(l)} = f(W \ast E^{(l-1)} + b)
\end{align}
% The encoder consists of $L$ layers $E^{(l)} \in \mathbb{R}^{F_{l} \times T_{l}}$, for $l \in \{0,\dots,L\}$, where $F_{l}$ is the number of latent states in a given layer and $T_{l}$ is the number of corresponding time steps. Each layer consists of three operations: compute a set of temporal convolutions, apply a non-linear activation per time-step, and perform max pooling across time. 
% % The convolutional filters aim to capture how the input signal changes over time. 
% The input features, $X$, are denoted $E^{(0)}$ for convenience.
% 
% Each convolution layer consists of $F_{l}$ filters, each of which is parameterized by weight matrix 
% % $W_i^{(l)} \in \mathbb{R}^{ F_{l-1} \times d}$ and bias $b_i^{(l)}  \in \mathbb{R}$
% $W^{(i)} \in \mathbb{R}^{ F_{l-1} \times d}$ and bias $b^{(i)}  \in \mathbb{R}$
% for filter index $i \in \{1,\dots,F_l\}$ and filter duration $d$. 
% Given the signal from the previous layer, $E^{(l-1)}$, the activations in the $l$-th layer are given by
% \begin{align}\label{eqn:ED-TCN}
% % E^{(l)}_{i,t} = f(\sum_{t'=0}^{d-1} \langle  W_{i,t',\cdot}^{(l)},  E^{(l-1)}_{\cdot,t-t'} \rangle + b_i^{(l)})\\
% % E^{(l)}_{i,t} = f(\sum_{t'=0}^{d-1} \langle  W_{t',\cdot}^{(i)},  E^{(l-1)}_{\cdot,t-t'} \rangle + b_i^{(l)})\\
% E^{(l)}_{\cdot,t} = f(\sum_{t'=0}^{d-1}  W^{(l)}_{t',\cdot} E^{(l-1)}_{\cdot,t-t'} + b^{(l)})\\
% % E^{(l)}_{i,t} = f( W_{i}^{(l)} *  E^{(l-1)}_{t-d+1:t} + b_i^{(l)})
% \end{align}
where $f(\cdot)$ is the activation function and  $\ast$ is the convolution operator. We compare activation functions in Section~\ref{sec:experiments} and find normalized Rectified Linear Units perform best.
After each activation function, we max pool with width 2 across time such that $T_l = \frac{1}{2} T_{l-1}$. 
Pooling enables us to efficiently compute activations over long temporal windows.
% Note that in theory this implies $T$ must be divisible by $2^L$. In practice, we pad each sequence to be of an appropriate length, given the pooling operations, such that the input length of the whole sequence, $T$, and the length of the output predictions are the same.

Our decoder is similar to the encoder, except that upsampling is used instead of pooling and the order of the operations is now upsample, convolve, and apply the activation function. 
Upsampling is performed by simply repeating each entry twice. The convolutional filters in the decoder distribute the activations from the condensed layers in the middle to the action predictions at the top.
Experimentally, these convolutions provide a large improvement in performance and appear to capture pairwise transitions between actions. 
Each decoder layer is denoted by 
% $D^{(l)} \in \mathbb{R}^{T_l \times F_{l}}$
$D^{(l)} \in \mathbb{R}^{F_{l} \times T_l}$
for $l \in \{L,\dots,1\}$. Note that these are indexed in reverse order compared to the encoder, so the filter count in the first encoder layer is the same as in the last decoder layer. 

%The encoder and decoder are connected with a set of fully connected weights that are shared across time with weight vector $W \in \mathbb{R}^{F_L \times F_L} $ and bias $c  \in \mathbb{R}^{F_L}$:
%\begin{align}
%D^{(L)}_{t} =  W  E^{(L)}_{t} + c.
%\end{align}
%A temporal fully connected layer connects the encoder and decoder using a is defined with using $F_L \times 1$ convolutions such that for every temporal region, there will be one input vector and one output vector each with size $F_L$. Weights are shared across time. 
%\TODO{Note that an (unshared) fully connected layer can capture differences between the start and end of a video, however, it requires that all videos be of the same total time $T$.}

%Each layer in the decoder is connected with its corresponding encoder layer via a set of skip edges.
%For each time step, the decoder score is a function of the encoder activations at that timestep, $E^{(l)}_t$, and decoder activations, $D^{(l)}_t$. The score for each layer of the decoder is the following with convolutional filters $U_i^{(\cdot)} \in \mathbb{R}^{d \times F_{(\cdot)}}$ and biases $c^{(\cdot)} \in \mathbb{R}^{F_{(\cdot)}}$
%\begin{align}
%D^{(l)}_{i,t} = \sum_{t'=1}^{d} U_{i,t}^{(l)}  D^{(l+1)}_{t+d-t'}+c_i^{(l)}\\
%%D^{(l)}_t = U_i^{(l)} \star  D^{(l-1)}_{t:t+d} + c_i\\
%\hat{D}^{(l)}_t = D^{(l)}_t + V E^{(l)}_t + e
%\end{align}
%%This is in contrast to SegNet which carries over indices from max pooling but not the corresponding activtions.

The probability that frame $t$ corresponds each of the C action classes is given by vector $\hat Y_t \in [0,1]^C$ using weight matrix $U \in \mathbb{R}^{C \times F_1}$ and bias $c \in \mathbb{R}^{C}$, such that
% The probabilities for all time steps,  given the the decoder output, $D^{(1)}$, is
% Given the the decoder output, $D^{(1)}$, this probability is
\begin{align}
\hat{Y}_t = \text{softmax}(U D^{(1)}_t + c).
% \hat{Y} = \text{softmax}(U D^{(1)} + c).
\end{align}

We explored other mechanisms, such as skip connections between layers, different patterns of convolutions, and other normalization schemes, however, the proposed model outperformed these alternatives and is arguably simpler. 
%is the . These helped at times and hurt in others. The aforementioned solution was superior in aggregate. 
Implementation details are described in Section~\ref{sec:details}.

\fakesubsection{Receptive Field}
The prediction at each frame is a function of a fixed-length period of time, which is given by
% The receptive field is defined as the region over which each prediction is computed.
% , and is in large part why our model outperforms many others. 
%Typical approaches using RNNs and CRFs 
the $r(d,L)=d(2^L-1) + 1$ for $L$ layers and duration $d$. 
% \TODO{modify comparison: 
% In our experiments on 50 Salads the receptive field will be on the order of 30 seconds, which is much larger than the effective receptive field of LSTM on the MERL dataset~\cite{singh_cvpr_2016_merl} and longer than the temporal receptive field of recent spatiotemporal CNNs~\cite{lea_eccv_2016,ng_cvpr_2015}.
% }

%Residual connections~\cite{resnet} are applied between encoder and decoder layers. \TODO{Explain}

%The TCN architecture is depicted in Figure~\ref{fig:unified_model}. In practice we use a depth of $L=3$ but in principle this can be arbitrarily large. 

%For each frame in our video experiments, the input, $X_t$, is the first fully connected layer computed in a spatial CNN trained solely on each dataset. We trained the model of~\cite{lea_eccv_2016}, except instead of using Motion History Images (MHI) as input to the CNN, we concatenate the following for image $I_t$ at frame $t$: $[I_t, I_{t-d}-I_t, I_{t+d}-I_t, I_{t-2d}-I_t, I_{t+2d}-I_t]$ for $d=0.5$ seconds. 
%In our experiments, these difference images -- which can be viewed as a simple type of attention mechanism -- tend to perform better than MHI or optical flow across these datasets. 
%Furthermore, for each time step, we perform channel-wise normalization before feeding it into the TCN. This helps with large environmental fluctuations, such as changes in lighting. 
%%In the results, we show performance using only the spatial model, the spatiotemporal model of Lea \etal~\cite{lea_eccv_2016}, and the TCN.

% ------------------------------------------------------
\begin{figure}
	\centering
	\includegraphics[width=\hsize]{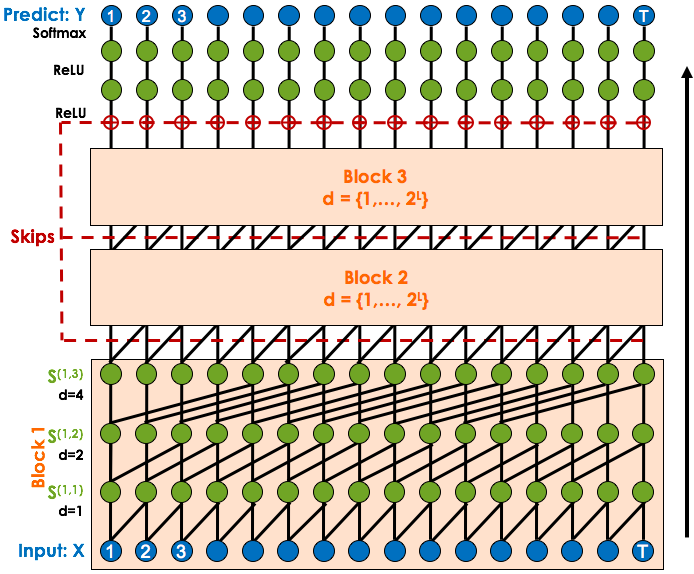}		    
	\caption{The Dilated TCN model uses a deep stack of dilated convolutions to capture long-range temporal patterns. }
	\label{fig:WaveNet}
\end{figure}

\subsection{Dilated TCN}

% We adapt the WaveNet model by van den Oord \etal~\cite{wavenet} for action segmentation. This is similar to our ED-TCN except that it uses hierarchies of dilated convolutions instead of pooling to capture long-range temporal patterns. 
We adapt the WaveNet~\cite{wavenet} model, which was designed for speech synthesis,  to the task of action segmentation. In their work, the predicted output, $Y_t$, denoted which audio sample should come next given the audio from frames $1$ to $t$.
% $X_1,\dots,X_{t}$.
In our case $Y_t$ is the current action given the video features up to $t$.
% $X_1,\dots,X_{t}$.

% \TODO{clean up notation}
As shown in Figure~\ref{fig:WaveNet}, we define a series of blocks, each of which contains a sequence of $L$ convolutional layers. The activations in the $l$-th layer and $j$-th block are given by $S^{(j,l)} \in \mathbb{R}^{F_{w} \times T}$.
% , for $l \in \{0,\dots,L\}$.
% where the input to a given block is $S^{(j,0)}$.
Note that each layer has the same number of filters $F_{w}$.
% , other than the input signal,
This enables us to combine  activations from different layers later. 
Each layer consists a set of dilated convolutions with rate parameter $s$, a non-linear activation $f(\cdot)$, and a residual connection than combines the layer's input and the convolution signal. 
Convolutions are only applied over two time steps, $t$ and $t-s$, so we write out the full equations to be specific.
% Therefore $W \in \mathbb{R}^{2 \times F_w \times F_w}$ with bias vector $b^{(b,l)}  \in \mathbb{R}^{F_{l=w}}$. 
The filters are parameterized by $W = \{W^{(1)}, W^{(2)}\}$ with $W^{(i)} \in \mathbb{R}^{ F_w \times F_w}$ and bias vector $b \in \mathbb{R}^{F_{w}}$. 
Let $\hat{S}^{(j,l)}_{t}$ be the result of the dilated convolution at time $t$ and $S^{(j,l)}_{t}$ be the result after adding the residual connection such that
% \begin{align}\label{eqn:WaveNet}
% \hat{S}^{(b,l)}_{t} &= f(W_{2} S^{(b,l-1)}_{t-s} + W_{1} S^{(b,l-1)}_{t} + b)\\
% S^{(b,l)}_{t} &= S^{(b,l-1)}_{t-s} + V \hat{S}^{(b,l)}_{t} + d.
% \end{align}
\begin{align}\label{eqn:WaveNet}
\hat{S}^{(j,l)}_{t} &= f(W^{(1)} S^{(j,l-1)}_{t-s} + W^{(2)} S^{(j,l-1)}_{t} + b)\\
S^{(j,l)}_{t} &= S^{(j,l-1)}_{t} + V \hat{S}^{(j,l)}_{t} + e.
% V^{(b,l)}
\end{align}
Let $V \in \mathbb{R}^{F_w \times F_w}$ and $e \in \mathbb{R}^{F_w}$ be a set of weights and biases for the residual. Note that parameters $\{W,b,V,e\}$ are separate for each layer.
% \TODO{thought: what if I remove superscripts from all weights?}

% These dilations enable the convolution operation to reach values beyond the immediate neighborhood around the value that is being evaluated.
The dilation rate increases for consecutive layers within a block such that $s_l=2^l$.
This enables us to increase the receptive field by a substantial amount without drastically increasing the number of parameters. 
% By exponentially increasing the dilation factor at each layer, we can obtain a much larger receptive field while only using small filters throughout the network. 
% \begin{align}\label{eqn:dilated-conv-layer}
% E^{(l)} = f(W^{(l)}\ast_n E^{(l-1)} + b^{(l)})
% \end{align}

% Their model is similar to the Multi-Scale Context CNN by Yu and Koltun~\cite{yu_iclr_2016} but applied to .
% A similar model was used by  for speech synthesis and music generation, which we draw inspiration from as well.

% For a discrete function $f:\mathbb{Z}\rightarrow\mathbb{R}$ and filter $g:\mathbb{Z}\rightarrow\mathbb{R}$ with indices in the set $[-T, T]$, the discrete convolution operation $\ast$ is defined as
% \begin{align}\label{eqn:convolution}
% (f\ast g)(t) = \sum_{\tau=-T}^T f(t-\tau)g(\tau)
% \end{align}=========

% Now let $\ast_n$ be a discrete dilated convolution operator with dilation $n$, defined as
% \begin{align}\label{eqn:dilated-conv}
% (f\ast_n g)(t) = \sum_{\tau=-T}^T f(t-n\tau)g(\tau)
% \end{align}

% We can implement a dilated convolution layer using the same notation before, albeit using a dilated convolution operator instead.
% \begin{align}\label{eqn:dilated-conv-layer}
% E^{(l)} = f(W^{(l)}\ast_n E^{(l-1)} + b^{(l)})
% \end{align}

The output of each block is summed using a set of skip connections with $Z^{(0)} \in \mathcal{R}^{T \times F_w}$ such that 
\begin{align}
Z_t^{(0)} = ReLU(\sum_{j=1}^B S_t^{(j,L)}).
\end{align}
There is a set of latent states $Z^{(1)}_t = ReLU(V_r Z^{(0)}_t + e_r)$ for weight matrix $V_r \in \mathbb{R}^{F_w \times F_w}$ and bias $e_r$. 
The predictions for each time $t$ are given by
\begin{align}
\hat{Y}_t = softmax(U Z^{(1)}_t + c)
\end{align}
for weight matrix $U \in \mathbb{R}^{C \times F_w}$ and bias $c \in \mathbb{R}^{C}$.

\fakesubsection{Receptive Field} The filters in each Dilated TCN layer are smaller than in ED-TCN, so in order to get an equal-sized receptive field it needs more layers or blocks. The expression for its receptive field is $r(B,L)= B*2^L$ for number of blocks $B$ and number of layers per block $L$.

\subsection{Implementation details}
\label{sec:details}
Parameters of both TCNs are learned using the categorical cross entropy loss with Stochastic Gradient Descent and ADAM~\cite{ADAM} step updates. 
Using dropout on full convolutional filters~\cite{tompson_cvpr_2015}, as opposed to individual weights, improves performance and produces smoother looking weights.
For ED-TCN, each of the $L$ layers has $F_l=96 + 32*l$ filters.
% except for the $F_0$ which is defined by the input features. 
For the Dilated TCN we find that performance does not depend heavily on the number of filters for each convolutional layer, so we set $F_w=128$.
We perform ablative analysis with various number of layers and filter durations in the experiments.
All models were implemented using Keras~\cite{keras} and TensorFlow~\cite{tensorflow}.

% As with the ED-TCN, we use a categorical cross entropy loss using SGD with ADAM. Unless otherwise stated we use $B=2$ blocks with $L=5$ layers. We 
%Filter duration, $d$, is set as the mean segment duration for the shortest class from the training set. For example, $d=10$ seconds for 50 Salads. 

\subsection{Causal versus Acausal}

We perform causal and acausal experiments. Causal means that the prediction at time $t$ is only a function of data from times $1$ to $t$, which is important for applications in robotics. Acausal means that the predictions may be a function of data at any time step in the sequence. 
% The models described are causal, meaning, they only use data captured at or before a given time step $t$. 
For the causal case in ED-TCN, for at each time step $t$ and filter length $d$, we convolve from $X_{t-d}$ to $X_t$. In the acausal case we convolve from $X_{t-\frac{d}{2}}$ to $X_{t+\frac{d}{2}}$. 
% However, for offline applications such as video summarization it may be advantageous to use the data before and after a given time step. As such, we experiment with acausal variations of each model. For ED-TCN we replace Eqn~\ref{eqn:ED-TCN} with 
% \begin{align}
% E^{(l)}_{t,i} = f(\sum_{t'=1}^{d}  W^{(i)}_{t'} E^{(l-1)}_{t+\frac{d}{2}-t'} + b^{(i)})
% \end{align}
% \begin{align}
% %E^{(l)}_{i,t} = f_{ED}( b_i^{(l)} + \sum_{t'=0}^{d-1} \langle  W_{i,t',\cdot}^{(l)},  E^{(l-1)}_{\cdot,t-t'} \rangle )
% E^{(l)}_{i,t} = f( W_{i}^{(l)} \star  E^{(l-1)}_{t-\frac{d}{2}:t+\frac{d}{2}} + b_i^{(l)})
% \end{align}

For the acausal Dilated TCN, we modify Eqn~\ref{eqn:WaveNet} such that each layer now operates over one previous step, the current step, and one future step:
% we replace Eqn~\ref{eqn:WaveNet} with 
\begin{align}
% \hat{S}^{(b,l)}_{t} = f(W^{(b,l)}_{3} S^{(b,l-1)}_{t-s} &+ W^{(b,l)}_{2} S^{(b,l-1)}_{t} \nonumber\\ 
% &+ W^{(b,l)}_{1} S^{(b,l-1)}_{t+s} + b^{(l)}).
\hat{S}^{(j,l)}_{t} = f(W^{(1)} S^{(j,l-1)}_{t-s} &+ W^{(2)} S^{(j,l-1)}_{t}  \nonumber\\ 
&+ W^{(3)} S^{(j,l-1)}_{t+s} + b)
\end{align}
where now $W = \{W^{(1)}, W^{(2)}, W^{(3)}\}$.

\section{Evaluation \& Discussion}
\label{sec:evaluation}
We start by performing synthetic experiments that highlight the ability of our TCNs to capture high-level temporal patterns. We then perform quantitative experiments on three challenging datasets and ablative analysis to measure the impact of hyper-parameters such as filter duration.

% We perform three types of experiments: 
% (1) synthetic experiments highlighting the capability of our model to capture higher-order temporal patterns,
% (2) quantitative action segmentation results, including ablative analysis of our model, and 
% (3) qualitative results visualizing parameters and outputs of our model. 

%------------------------------------------------------------------------------------
\subsection{Metrics}
Papers addressing action segmentation tend to use different metrics than those on action detection. We evaluate using metrics from both communities and introduce a segmental F1 score, which is applicable to both tasks.
% intended to bridge the gap between evaluation methods and overcome fundamental limitations with the current metrics. 

\fakesubsection{Segmentation metrics}
Action segmentation papers tend use to frame-wise metrics, such as accuracy, precision, and recall~\cite{stein_ubicomp_2013,kuehne_cvpr_2014}. Some work on 50 Salads also uses segmental metrics~\cite{lea_eccv_2016,lea_icra_2016}, in the form of a segmental edit distance, which is useful because it penalizes predictions that are out-of-order and for over-segmentation errors. We evaluate all methods using frame-wise accuracy.

One drawback of frame-wise metrics is that models achieving similar accuracy may have large qualitative differences, as visualized later.
For example, a model may achieve high accuracy but produce numerous over-segmentation errors. It is important to avoid these errors for human-robot interaction and video summarization. 
% For example, on 50 Salads achieve nearly the same performance, but two of them tend to introduce notably more over-segmentation errors than the other. This will be apparent when we visualize the predictions later. Thee

\fakesubsection{Detection metrics}
Action detection papers tend to use segment-wise metrics such as mean Average Precision with midpoint hit criterion (mAP@mid)~\cite{rohrbach_ijcv_2015,singh_cvpr_2016_merl} or mAP with a intersection over union (IoU) overlap criterion (mAP@k)~\cite{richard_cvpr_2016}. 
mAP@k is computed my comparing the overlap score for each segment with respect to the ground truth action of the same class. If an IoU score is above a threshold $\tau=\frac{k}{100}$ it is considered a true positive otherwise it is a false positive. Average precision is computed for each class and the results are averaged. 
mAP@mid is similar except the criterion for a true positive is whether or not the midpoint (mean time) is within the start and stop time of the corresponding correct action.

% These mAP metrics were adapted from the object detection literature and are good at evaluating retrieval tasks. To compute these metrics, action predictions are ranked such that higher confidence predictions are scores before lower-confidence predictions. This makes sense when you are displaying several different predictions to a user -- as is typical for retrieval tasks -- but is not a useful metric for applications these datasets were designed around.  For example, in video summarization, the order in which actions were performed is of great importance, so the segmental edit score is useful. 
% For surveillance tasks, it may be useful to display a timeline summary video to a user. In this case, 

mAP is a useful metric for information retrieval tasks like video search, however for many fine-grained action detection applications, such as robotics or video surveillance, we find that results are not indicative of real-world performance. 
% To understand this limitation, One particularly insightful limitation is as follows. 
The key issue is that mAP is very sensitive to a confidence score assigned to each segment prediction. 
% , and the final score is a function of this confidence. This is useful for applications  In 
% To compute mAP scores, each of the action detections is ranked using a set of confidence scores. 
These confidences are often simply the mean or maximum class score within the frames corresponding to a predicted segment.
% While mAP metrics are commonly used for action detection datasets (e.g.,~\cite{rohrbach_ijcv_2015,singh_cvpr_2016_merl}), 
By computing these confidences in subtly different ways you obtain wildly different results. 
On MERL Shopping, the mAP@mid scores for Singh \etal~\cite{singh_cvpr_2016_merl} jump from 50.9 using the mean prediction score over an interval to 69.8 using the maximum score over that same interval. 

\fakesubsection{F1@k}
We propose a segmental F1 score which is applicable to both segmentation and detection tasks and has the following properties:
(1) it penalizes over-segmentation errors, (2) it does not penalize for minor temporal shifts between the predictions and ground truth, which may have been caused by annotator variability, and (3) scores are dependent on the number actions and not on the duration of each action instance. 
This metric is similar to mAP with IoU thresholds except that it does not require a confidence for each prediction. 
Qualitatively, we find these numbers are better at indicating the caliber of a given segmentation. 

We compute whether or not each predicted action segment is a true or false positive by comparing its IoU with respect to the corresponding ground truth using threshold $\tau$. 
As with mAP detection scores, if there is more than one correct detection within the span of a single true action then only one is marked as a true positive and all others are false positives. We compute precision and recall for true positives, false positives, and false negatives summed over all classes and compute 
% an F1 score, which is given by 
$F1=2\frac{prec*recall}{prec+recall}$. 
% We evaluate using three overlap values, denoted by $F1@k$, where $k$ is the overlap threshold percentage. 

We attempted to obtain action predictions from the original authors on all datasets to compare across multiple metrics. We received them for 50 Salads and MERL Shopping.

% This edit score is computed by applying the Levenshtein distance to the segmented predictions (e.g. $AAABBA \rightarrow ABA$). This is normalized to be in the range $0$ to $100$ such that higher is better. }

% In the supplemental material we provide an exhaustive listing of metrics including frame-wise accuracy, mAP (midpoint) ... 

\subsection{Synthetic Experiments}
\begin{figure}
	\center
	\includegraphics[width=\hsize]{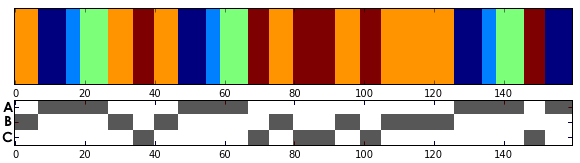}
	\caption{Synthetic Experiment \#1: (top) True action labels for a given sequence (bottom) The 3 dimensional features for that sequence. White is $-1$ and gray is $+1$. Subctions A1, A2, and A3 (dark blue, light blue and green) all have to the same feature values, which differ from B (orange) and C (red).}
	\label{fig:composites}
\end{figure}

We claim TCNs are capable of capturing complex temporal patterns, such as action compositions, action durations, and long-range temporal dependencies. We show these abilities with two synthetic experiments. 
% The first highlights the ability to learn temporal shifts and 
% The first demonstrates the ability to model composite actions and the duration of actions.
For each, we generate toy features $X$ and corresponding labels $Y$ for 50 training sequences and 10 test sequences of length $T=150$. The duration of each action of a given class is fixed and action segments are sampled randomly. An example for the composition experiment is shown in Figure~\ref{fig:composites}.
Both TCNs are acausal and have a receptive field of length 16.
% The action segments were generated using a Markov model with three action classes ($C=3$).
% We randomly sampled segments with fixed duration
% For each experiment we generated 50 sequences of length $T=150$. 
% The values of each feature are $+1$ for the true class at that time and $-1$ for the incorrect classes. We randomly sampled segments with fixed duration (e.g., segments corresponding to class 1 have duration 6, for class 2 have duration 13, and 3 have duration 17). 

\fakesubsection{Action Compositions}
By definition, an activity is composed of a sequence of actions. Typically there is a dependency between consecutive actions (e.g., action $B$ likely comes after $A$).
CRFs capture this using a pairwise transition model over class labels and RNNs capture it using LSTM across latent states. 
We show that TCNs can capture action compositions, despite not explicitly conditioning the activations at time $t$ on previous time steps within that layer. 

% Despite the fact that TCNs do not explicitly model the relationships between the previous label, $Y_{t-1}$, and the current label, $Y_t$, we show that in practice they are capable of ...\TODO{rewrite}
% One benefit of CRFs and RNNs is that they typically model the relationship between multiple sequential actions. CRFs, for example, encode the probability of action $B$ happening after action $A$. 
% As demonstrated in~\cite{richard_cvpr_2016}, they can also model the durations of action segments. 
% In this experiment we highlight 
% Actions can be defined hierarchically. For example, the high-level action \textit{cutting} may consist of three low-level sub-actions \textit{pick up knife}, \textit{cut tomato}, and \textit{place knife down}. In this experiment we demonstrate that ED-TCN and Dilated TCN can capture of composite actions even if they are not easy to distinguish from low-level features.

In this experiment, we generated sequences using a Markov model with three high-level actions $A$, $B$, and $C$ with subactions $A_1$, $A_2$, and $A_3$, as shown in Figure~\ref{fig:composites}. $A$ always consists of subactions $A_1$ (dark blue), $A_2$ (light blue), then $A_3$ (green), after which it is transitions to $B$ or $C$.
% (orange) (red)
% Let $A$ can transition to $B$ or $C$, $B$ can transition to $A$ or $C$, and $C$ can transition to $A$ or $B$. 
% We generated sequences using a grammar with actions $a_1$, $a_2$, $a_3$, $b$, and $c$. Our grammar $R=\{S=>A|B|C, A=>AB|AC, B=>bA|bC, C=>cA|cB, A=>A_1A_2A_3, B=>b, C=>c\}$. 
For simplicity, $X_t \in \mathbb{R}^3$ corresponds to the high-level action $Y_t$ such that the true class is $+1$ and others are $-1$. 
% The features for $A_1$-$A_3$ are the same, which are different from B and C. Figure~\ref{fig:composites} shows example features and labels.

% \TODO{change notation} We generated sequences using a context free grammar with three high-level actions ($A$, $B$, and $C$) and subactions ($a_1$, $a_2$, and $a_3$). We use the grammar $R=\{S=>A|B|C, A=>AB|AC, B=>bA|bC, C=>cA|cB, A=>A_1A_2A_3, B=>b, C=>c\}$ to create our sequences. The features for A1-A3 are the same, which are different from B and C. Figure~\ref{fig:composites} shows example features and labels.

The feature vectors corresponding to subactions $A1-A3$ are all the same, thus a simple frame-based classifier would not be able to differentiate them. The TCNs, given their long receptive fields, segment the actions perfectly. 
% A simple frame-based classifier would not be able to differentiate between subactions $A1-A3$ because they have the same feature feature vectors, however, both TCNs achieve 100\% accuracy. 
This suggests that our TCNs are capable of capturing action compositions. 
% between the subactions.
% , despite not having knowledge that each action is occurring. 
Recall each action class had a different segment duration, and we correctly labeled all frames, which suggests TCNs can capture duration properties for each class. 
% class at  so it is importance to note that these models are capable of capturing this duration, as well as the fact that the subactions transitions in order from $A_1$ through $A_3$.
% Our LSTM-based model is capable of capturing the pairwise transitions between subactions, but does not accurately capture the duration of each action. 
% It is important to note that the duration of each action class is distinct. Therefore, these models not only learn the sequencing of actions (e.g., class B proceeds A) but also captures the duration of each action. 
% \TODO{[Insert Bi-LSTM accuracy]}
% We also see this visually in our qualitative experiments. 

\begin{table}
	\centering
	\begin{tabular}{|l|c|c|c|c|c|}
		\hline
		Shift & s=0 & s=5 & s=10 & s=15 & s=20\\
		\hline		
% 		ED-TCN  & 0/100 & 0/97.9 & 0/89.5 & 1/74.1 & 9/57.1 \\
% 		Dilated TCN & 0/100 & 0/92.7 & 0/87.0 & 0/69.6 & 0/61.5\\
% 		Bi-LSTM & 0/100 & 5/72.3 & 10/60.2 & 12/54.7 & 15/38.5\\   
% ED-TCN  & 0/100 & 0/97 & 0/89 & 1/74 & 9/57\\
% Dilated TCN & 0/100 & 0/92 & 0/87 & 0/69 & 0/61\\
% Bi-LSTM & 0/100 & 5/72 & 10/60 & 12/54 & 15/38\\
ED-TCN  & 100 & 97.9 & 89.5 & 74.1 & 57.1 \\
Dilated TCN & 100 & 92.7 & 87.0 & 69.6 & 61.5\\
Bi-LSTM & 100 & 72.3 &  60.2 & 54.7 & 38.5\\ 
		\hline
	\end{tabular}
	\label{tab:shift_results}	
	\caption{Synthetic experiment \#2: F1@10 when shifting the input features in time with respect to the true labels. 
%     $X/Y$ indicates the median delayed detection rate between the true start time for each action and the production and the segmental F1 score (overlap=0.1).
    Column shows performance when shifting the data $s$ frames from the corresponding labels. Each TCN receptive field is 16 frames.}
\end{table}

\fakesubsection{Long-range temporal dependencies}
For many actions it is important to consider information from seconds or even minutes in the past.
% TCNs are capable of capturing mid-range temporal dependencies such as the change in state several seconds ago. 
% There are many cases in action segmentation where the salient motions that indicate an action is happening do not occur at the same time as the actions themselves. 
For example, in the cooking scenario, when a user cuts a tomato, they tend to occlude the tomato with their hands, which makes it difficult to identify which object is being cut. It would be advantageous to recognize that the tomato is on the cutting board before the user starts the cutting action. 
In this experiment, we show TCNs are capable of learning these long-range temporal patterns by adding a phase delay to the features. Specifically, for both training and test features we define $\hat{X}$ as $\hat{X}_t=X_{t-s}$ for all $t$. Thus, there is a delay of $s$ frames between the labels and the corresponding features. 

Results using F1@10 are shown in~\ref{tab:shift_results}. For comparison we show the TCNs as well as Bi-LSTM.
As expected, with no delay ($s=0$) all models achieve perfect prediction. For short delays ($s=5$), TCNs correctly detect all actions except the first and last of a sequence. 
As the delay increases, ED-TCN and Dilated TCN perform very well up to about half the length of the receptive field. 
The results for Bi-LSTM degrade at a much faster rate.

% We take the synthesized data described above and artificially shift the feature vectors by a fixed time $s$. 
% Specifically, the features at $t$ now correspond to the labels at $t-s$.  

% We assess performance using median delayed detection rate and F1@10. The delay rate refers to median difference between the true start time of each segment and the predicted start time. Results are shown in~\ref{tab:shift_results}. As expected, with no delay ($s=0$) both models achieve perfect prediction. 
% As the delay increases, ED-TCN and Dilated TCN perform very well up to about half the length of the receptive field. For comparison we also evaluated using Bi-LSTM and the results degrade at a much faster rate than either TCN. 

% ------------------------------------------

% ---------------------------------------
\subsection{Datasets}

% We evaluate on three challenging public datasets and compare performance with other models.

% \subsubsection{Datasets}

% We evaluate on three challenging public datasets. \TODO{mention lack of data from MPII, etc?}.
%The input signal, $X_t$, for the video results are the same as extracted from the spatial component of the spatiotemporal CNN in~\cite{lea_eccv_2016}.

%There are many questions that can be asked of this system. 
%We perform a comprehensive set of experiments to show ablative analysis, the effect of training on different action granularities,
% transfer of learned features from one granularity (e.g. very fine) to another (e.g. course), 
% and how our model performs if we assume temporal segmentation. Note that our emphasis is not on how the accuracy alone improves, but on how well the segments are modeled. For many applications having a set of cohesive segments is at least, if not more, important than good accuracy. 

%We perform a set of experiments: (1) How does each component of our model (spatial, spatial-temporal, segmental) contribute to overall performance? (2) How does using finer- versus coarser-grained action labels impact performance? (3) Does performance change if we train with a set of finer-grained set of labels than we need? (4) Can we train on one fine-grained dataset and test on another in a different domain? (5) How does our model perform if we have known temporal segmentation? \TODO{say this in words}

%\TODO{reference ICRA paper}

\fakesubsection{University of Dundee 50 Salads~\cite{stein_ubicomp_2013}}
contains 50 sequences of users making a salad and has been used for both action segmentation and detection. 
While this is a multimodal dataset we only evaluate using the video data. 
% with multiple sensor types
Each video is 5-10 minutes in duration and contains around 30 instances of actions such as \texttt{cut tomato} or \texttt{peel cucumber}. 
% This dataset includes video and synchronized accelerators attached to ten objects in the scene, such as the \textit{bowl}, \textit{knife}, and \textit{plate}. 
% here and in the supplemental material evaluate using the sensor data. 
We performed cross validation with 5 splits on a higher-level action granularity which includes 9 action classes such as \texttt{add dressing}, \texttt{add oil}, \texttt{cut}, \texttt{mix ingredients}, \texttt{peel}, and \texttt{place}, plus a background class. In ~\cite{stein_ubicomp_2013} this was referred to as ``eval.''
We also evaluate on a mid-level action granularity with 17 action classes. 
% The higher-level label set contains 9 action classes plus a background class and the mid-level contains 17 classes. 
The mid-level labels differentiates actions like \texttt{cut tomato} from \texttt{cut cucumber} whereas the higher-level combines these into a single class, \textit{cut}. 

% It is reasonable to assume we can recognize these actions using video or accelerometer information. 

%Our sensor results used the features from \cite{lea_icra_2016} which are the absolute values of accelerometer values. 
%Previous results (e.g., \cite{lea_icra_2016,richard_cvpr_2016}) were evaluated using different setups. For example, \cite{lea_icra_2016} smoothed out short interstitial background segments. We reran all results to be consistent with~\cite{richard_cvpr_2016}. We also included an LSTM baseline for comparison which uses $64$ hidden states. 
We use the spatial CNN features of Lea \etal~\cite{lea_eccv_2016} as input into our models. This is a simplified VGG-style model trained solely on 50 Salads. 
% These use RGB and difference images as input into a simplified VGG-style CNN. 
Data was downsampled to approximately 1 frame/second. 
% We evaluate using the splits described in~\cite{stein_ubicomp_2013}. 
% A subset of the training data is used as validation for optimizing hyper parameters. 

\fakesubsection{MERL Shopping~\cite{singh_cvpr_2016_merl}} is an action detection dataset
consisting of 106 surveillance-style videos in which users interact with items on store shelves. The camera viewpoint is fixed and only one user is present in each video. There are five actions plus a background class: \texttt{reach to shelf},  \texttt{retract hand from shelf}, \texttt{hand in shelf}, \texttt{inspect product}, \texttt{inspect shelf}. Actions are typically a few seconds long. 

We use the features from Singh \etal~\cite{singh_cvpr_2016_merl} as input. Singh's model consists of four VGG-style spatial CNNs: one for RGB, one for optical flow, and ones for cropped versions of RGB and optical flow. We stack the four feature-types for each frame and use Principal Components Analysis with 50 components to reduce the dimensionality.
The train, validation, and test splits are the same as described in~\cite{singh_cvpr_2016_merl}.
Data is sampled at 2.5 frames/second.

\fakesubsection{Georgia Tech Egocentric Activities (GTEA)~\cite{fathi_cvpr_2011}}
contains 28 videos of 7 kitchen activities such as making a sandwich and making coffee. The four subjects performed each activity once.
The camera is mounted on the user's head and is pointing towards their hands.
On average there are about 19 (non-background) actions per video and videos are around a minute long. We use the 11 action classes defined in~\cite{fathi_iccv_2011} and evaluate using leave one user out. Cross-validation is performed over users 1-3 as done by~\cite{singh_cvpr_2016_ego}. 
We use a frame rate of 3 frames per second.
% We show results for user 2 to be consistent with \cite{fathi_iccv_2011} and \cite{singh_cvpr_2016_ego}.

We were unable to obtain state of the art features from \cite{singh_cvpr_2016_ego}, so 
we trained spatial CNNs from scratch using code from~\cite{lea_eccv_2016}, which was originally applied to 50 Salads. 
This is a simplified VGG-style network where the input for each frame is a pair of RGB and motion images. 
Optical flow is very noisy due to large amounts of video compression in this dataset, so we simply use difference images, such that for image $I_t$ at frame $t$ the motion image is the concatenation of $[I_{t-d}-I_t, I_{t+d}-I_t, I_{t-2d}-I_t, I_{t+2d}-I_t]$ for delay $d=0.5$ seconds. 
These difference images can be viewed as a simple attention mechanism.
We compare results from this spatial CNN, the spatiotemporal CNN from~\cite{lea_eccv_2016}, EgoNet~\cite{singh_cvpr_2016_ego}, and our TCNs. 

\subsection{Experimental Results}
\label{sec:experiments}

To make the baselines more competitive, we apply Bidirectional LSTM (Bi-LSTM)~\cite{bi_lstm} to 50 Salads and GTEA. We use 64 latent states per LSTM direction with the same loss and learning methods as previously described. The input to this model is the same as for the TCNs. Note that MERL Shopping already has this baseline. 

% \subsubsection{Action Segmentation}

% First we show results on each of the aforementioned datasets and then we show ablative results, including a comparison using different activation functions in our TCNs, using different sizes of receptive fields, and using causal versus acausal convolutions. Unless otherwise stated we use acausal convolutions. 
% ED-TCN has $L=3$ with $d=18$ and Dilated TCN has $B=2$ with $L=5$.

\begin{table}
\centering
%     \begin{tabular}{cc}
		%		 &  \\
		% -----------Salads----------
		\begin{tabular}{| l | c | c | c |}
			\hline
			\textbf{50 Salads (higher)} & \textbf{F1@$\{10,25,50\}$} & \textbf{Edit} & \textbf{Acc} \\
			\hline
			%sCNN (framewise)  & 24.10 & 66.64\\ 
			%stCNN (framewise)  & 49.6 & 69.35\\ 
% 			VGG \cite{lea_eccv_2016}  & & 7.6			& 38.3 \\           
% 			IDT \cite{lea_eccv_2016}  & & 16.8 & 54.3			\\
% 			Seg-ST-CNN \cite{lea_eccv_2016}  & \textbf{62.0} & 72.0\\ 
			%           TCN  & \textbf{65.1} & \textbf{72.4}\\\
			%           \hline
			Spatial CNN~\cite{lea_eccv_2016} & 35.0, 30.5, 22.7 &  25.5  & 68.0 \\
            Dilated TCN & 55.8, 52.3, 44.3 & 46.9 & 71.1 \\
			ST-CNN~\cite{lea_eccv_2016} & 61.7, 57.3, 47.2 & 52.8 & 71.3 \\
			Bi-LSTM & 72.2, 68.4, 57.8 & 67.7 & 70.9 \\            
            
ED-TCN & \textbf{76.5}, \textbf{73.8}, \textbf{64.5} & \textbf{72.2} & \textbf{73.4}\\            
			%           TCN (ours) & \textbf{78.5} & \textbf{83.7}\\            
			\hline
            \textbf{50 Salads (mid)} & \textbf{F1@$\{10,25,50\}$} & \textbf{Edit} & \textbf{Acc} \\
            \hline
            Spatial CNN~\cite{lea_eccv_2016} & 32.3, 27.1, 18.9 & 24.8 & 54.9 \\
            IDT+LM~\cite{richard_cvpr_2016} & 44.4, 38.9, 27.8 & 45.8 & 48.7\\
            Dilated TCN  & 52.2, 47.6, 37.4 & 43.1 & 59.3 \\
ST-CNN \cite{lea_eccv_2016} & 55.9, 49.6, 37.1 & 45.9 & 59.4\\
            Bi-LSTM & 62.6, 58.3, 47.0 & 55.6 & 55.7 \\ 
%             ED-TCN & \textbf{64.9}, \textbf{58.4}, \textbf{45.9} & \textbf{59.4} & 58.4 \\ 
ED-TCN & \textbf{68.0}, \textbf{63.9}, \textbf{52.6} & \textbf{59.8} & \textbf{64.7} \\
			\hline            
		\end{tabular}
        \label{tab:50salads}
        \caption{Action segmentation results 50 salads.
        F1@k is our segmental F1 score, Edit is the Segmental edit score from~\cite{lea_icra_2016}, and Acc is frame-wise accuracy. 
        }
\end{table}

\begin{table}
\centering 
\begin{tabular}{| l | c  | c | c |}
\hline
\textbf{MERL (acausal)} &  \textbf{F1@$\{10,25,50\}$} & \textbf{mAP} & \textbf{Acc} \\
\hline
MSN Det \cite{singh_cvpr_2016_merl}  & 46.4,  42.6, 25.6  & \textbf{81.9} & 64.6 \\
MSN Seg \cite{singh_cvpr_2016_merl} &  80.0, 78.3, 65.4 & 69.8& 76.3 \\
Dilated TCN & 79.9, 78.0, 67.5& 75.6 & 76.4  \\
ED-TCN &  \textbf{86.7}, \textbf{85.1}, \textbf{72.9}& 74.4 & \textbf{79.0}    \\      
\hline
\textbf{MERL (causal)} &  \textbf{F1@$\{10,25,50\}$} & \textbf{mAP} & \textbf{Acc} \\
\hline
MSN Det  \cite{singh_cvpr_2016_merl} & - & \textbf{77.6} & -  \\ 
Dilated TCN  & 72.7, 70.6, 56.5 & 72.2 & 73.0  \\
ED-TCN  & \textbf{82.1, 79.8, 64.0}  & 64.2 & \textbf{74.1}\\
\hline
\end{tabular}
\label{tab:MERL}
\caption{MERL Shopping results. 
% $\Leftrightarrow$ means Acausal and $\Rightarrow$ means causal. 
% F1@k is the segmental F1 score with overlap k\%. ED-TCN: conv=3, [32,64], 100 iter
Action segmentation results on the MERL Shopping dataset. Causal only uses features from previous time steps and acausal uses previous and future time steps. mAP refers to mean Average Precision with midpoint hit criterion.
}
\end{table}

\begin{table}
\centering
\begin{tabular}{| l | c | c  |}
% \multicolumn{3}{c}{GTEA}\\
\hline
\textbf{GTEA} & \textbf{F1@\{10,25,50\}}   & \textbf{Acc} \\
\hline
EgoNet+TDD \cite{singh_cvpr_2016_ego}& - &  \textbf{64.4} \\
Spatial CNN~\cite{lea_eccv_2016}  & 41.8, 36.0, 25.1  &   54.1 \\
ST-CNN~\cite{lea_eccv_2016}  & 58.7, 54.4, 41.9 &  60.6 \\ 
Dilated TCN  & 58.8, 52.2, 42.2 &  58.3 \\ 
Bi-LSTM  & 66.5, 59.0, 43.6 &  55.5 \\ 
ED-TCN & \textbf{72.2}, \textbf{69.3}, \textbf{56.0} & 64.0  \\
\hline
\end{tabular}      
\label{tab:gtea}
\caption{Action segmentation results on the Georgia Tech Egocentric Activities dataset. F1@k is our segmental F1 score and Acc is frame-wise accuracy. 
% , Seg assumes known we know the start and end times for each segment, and 
}
\end{table}
% ---------------------------
\fakesubsection{50 Salads}
Results on both action granularities are included in Table~\ref{tab:50salads}. All methods are evaluated in acausal mode. 
ED-TCN outperforms all other models on both granularities and on all metrics. 
We also compare against Richard \etal~\cite{richard_cvpr_2016} who evaluated on the mid-level and reported using IoU mAP detection metrics. Their approach achieved 37.9 mAP@10 and 22.9 mAP@50. The ED-TCN achieves 64.9 mAP@10 and 42.3 mAP@50 and Dilated TCN achieves 53.3 mAP@10 and 29.2 mAP@50. 
% We believe the biggest difference between our results is that they use Dense Trajectories and we use CNN features.

\begin{figure*}
	\center
		\begin{tabular}{ c  c  c }
% \hspace{.1\hsize}	\textbf{50 Salads} \hspace{.1\hsize} & \hspace{.1\hsize} \textbf{MERL Shopping} \hspace{.10\hsize} & \hspace{.1\hsize} \textbf{GTEA} \hspace{.12\hsize}
\hspace{.13\hsize}	\textbf{50 Salads} \hspace{.08\hsize} & \hspace{.08\hsize} \textbf{MERL Shopping} \hspace{.08\hsize} & \hspace{.08\hsize} \textbf{GTEA} \hspace{.17\hsize}
\end{tabular}    
   	\includegraphics[width=.9\hsize]{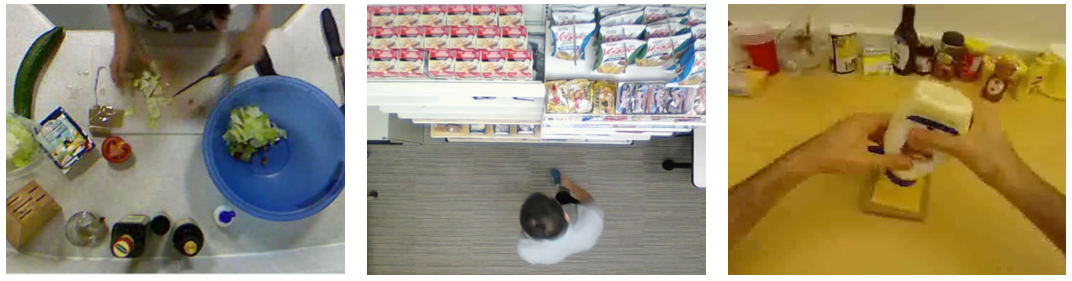}
	\includegraphics[width=.95\hsize]{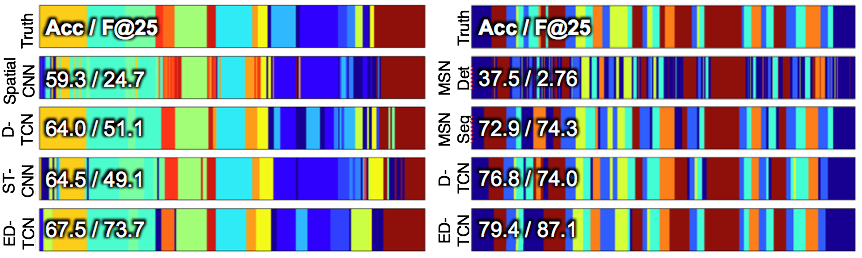}   
	\linebreak

	\caption{(top) Example images from each dataset. (bottom) Action predictions for one sequence using the mid-level action set of 50 Salads (left) and on MERL Shopping (right). These timelines are ``typical." Performance is near the average performance across each dataset.}
	\label{fig:timelines}
\end{figure*}

Notice that ED-TCN, Dilated TCN, and ST-CNN all achieve similar frame-wise accuracy but very different F1@k and edit scores. ED-TCN tends to produce many fewer over-segmentations than competing methods. Figure~\ref{fig:timelines} shows mid-level predictions for these models. Accuracy and F1 for each prediction is included for comparison.

Many errors on this dataset are due to the extreme similarity between actions and subtle differences in object appearance. For example, our models confuse actions using the vinegar and olive oil bottles, which have a similar appearance.
Similarly, we confuse some cutting actions (e.g., \texttt{cut cucumber} versus \texttt{cut tomato}) and placing actions (e.g., \texttt{place cheese} versus \texttt{place lettuce}).
% and the placing actions (e.g., \texttt{place cheese}, cucumber, lettuce, and tomato.

% ED-TCN achieves 59.7\% accuracy, 47.3\% edit, 
%  Richard: 54.2\% accuracy, edit score of 44.8\% edit, 

\fakesubsection{MERL Shopping}
We compare against use two sets of predictions from Singh \etal~\cite{singh_cvpr_2016_merl}, as shown in Table~\ref{tab:MERL}. The first, as reported in their paper, are a sparse set of action detections which are referred to as MSN Det. The second, obtained from the authors, are a set of dense (per-frame) action segmentations. The detections use activations from the dense segmentations with a non-maximal suppression detection algorithm to output a sparse set of segments. Their causal version uses LSTM on the dense activations and their acausal version uses Bidirectional LSTM. 

While Singh's detections achieve very high midpoint mAP, the same predictions perform very poorly on the other metrics. As visualized in Figure~\ref{fig:timelines} (right), the actions are very short and sparse. This is advantageous when optimizing for midpoint mAP, because performance only depends on the midpoint of a given action, however, it it not effective if you require the start and stop time of an activity. 
Interesting, this method does worst in F1 even for low overlaps. 

As expected the acausal TCNs perform much better than the causal variants. This verifies that using future information is important for achieving best performance. In the causal and acausal results the Dilated TCN outperforms ED-TCN in midpoint mAP, however, the F1 scores are better for ED-TCN. This suggests the confidences for the Dilated TCN are more reliable than ED-TCN. 

\begin{figure}
	\center
	\includegraphics[width=.48\hsize]{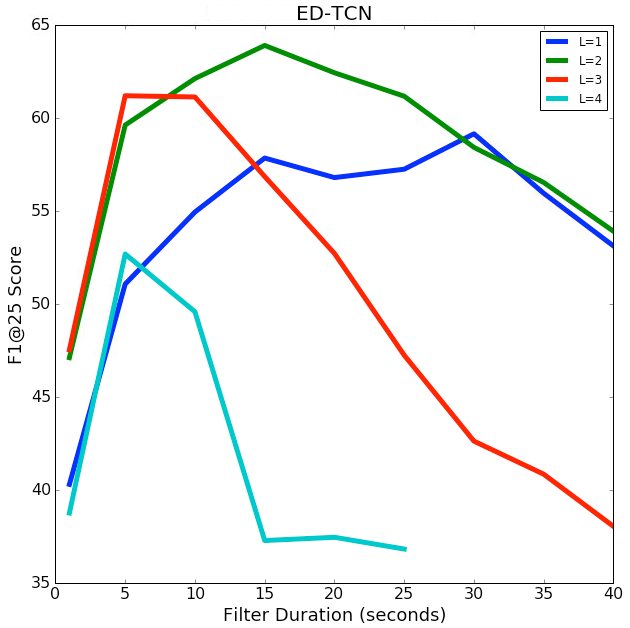} 
	\includegraphics[width=.48\hsize]{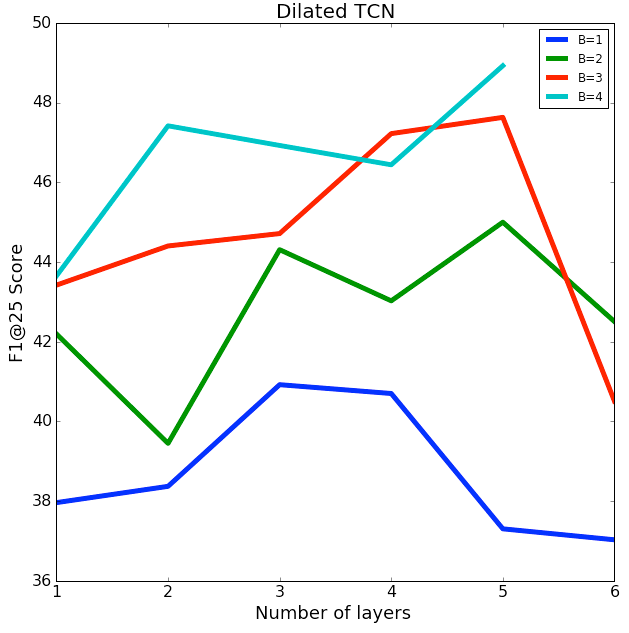}
	\caption{Receptive field experiments (left) ED-TCN: varying layer count $L$ and filter durations $d$  (right) Dilated TCN: varying layer count $L$ and  number of blocks $B$.}
	\label{fig:receptive_field}
\end{figure}

% \fakesubsection{Breakfast Actions}
\fakesubsection{Georgia Tech Egocentric Activities}
Performance of the ED-TCN is on par with the ensemble approach of Singh \etal~\cite{singh_cvpr_2016_ego}, which combines EgoNet features with TDD~\cite{TDD}. 
Recall that Singh's approach does not incorporate a temporal model, so we expect that combining their features with our ED-TCN would improve performance. 
Unlike EgoNet and TDD, our approach uses simpler spatial CNN features which can be computed in real-time.

Overall, in our experiments the Encoder-Decoder TCN outperformed all other models, including state of the art approaches for most datasets and our adaptation of the recent WaveNet model. 
The most important difference between these models is that ED-TCN uses fewer layers but has longer convolutional filters whereas the Dilated TCN has more layers but with shorter filters. 
The long filters in ED-TCN have a strong positive affect on F1 performance, in particular because they prevent over-segmentation issues. 
The Dilated TCN performs well on metrics like accuracy, but is less robust to over-segmentation. This is likely due to the short filter lengths in each layer.

% -------------------------------------
\subsubsection{Ablative Experiments}
Ablative experiments were performed on 50 Salads (mid-level). Note that these were done with different hyper-parameters and thus may not match previous results. 

\begin{table}
	\centering
	\begin{tabular}{|l|c|c|c|c|c|}
		\hline
		Activation  & Sigm. &  ReLU & Tanh   & GPC & NReLU  \\
		\hline
		ED-TCN  &  37.3 & 40.4 & 48.1 &   52.7 & \textbf{58.4} \\
		Dilated TCN & 42.5 &\textbf{43.1} &41.0  & 40.5 & 40.7 \\
		\hline
	\end{tabular}
	\label{tab:activations}	
	\caption{Comparison of different activation functions used in each TCN. Results are computed on 50 Salads (mid-level) with F1@25.}
\end{table}

\fakesubsection{Activation functions}
We assess performance using the activation functions shown in Table~\ref{tab:activations}.
% Sigmoid, Tanh, and ReLU are defined as in the literature.
The Gated PixelCNN (GPC) activation~\cite{GatedPixelCNN}, $f(x)=tanh(x) \odot sigmoid(x)$, was used for WaveNet and also achieves high performance on our tasks. 
We define the Normalized ReLU 
% as (NReLU)
\begin{align}
f(x) = \frac{ReLU(x)}{\max(ReLU(x))+\epsilon},
\end{align}
for vector $x$ and $\epsilon=1\text{\sc{e}-}5$ where the max is computed per-frame.
Normalized ReLU outperforms all others with ED-TCN, whereas for Dilated TCN all functions are similar. 
% The Gated PixelCNN and Normalized ReLU activations tend to achieve the best performance across experiments. 
% Computationally the Normalized ReLU is faster
% We find this achieves far superior performance to sigmoid, several percentage better than $tanh$ and $ReLU$, and is approximately the same as the WaveNet activation described later but more computationally efficient.

\fakesubsection{Receptive fields}
We compare performance with varying receptive field hyperparameters. 
Line in Figure~\ref{fig:receptive_field} (left) show F1@25 for $L$ from $1$ to $5$ and filter sizes $d$ from $1$ to $40$ on ED-TCN. 
Lines in Figure~\ref{fig:receptive_field} (right) correspond to block count $B$ with layer count $L$ from $1$ to $6$ for a Dilated TCN. Note, our GPU ran out of memory on ED-TCN after ($L=4$,$d=25$) and Dilated TCN after ($B=4$,$L=5$).
The ED-TCN performs best with a receptive field of 44 frames ($L=2$,$d=15$) which corresponds to 52 seconds. The Dilated TCN performs best at 128 frames ($B=4$,$L=5$) and achieves similar performance at 96 frames ($B=3$,$L=5$). 
% Larger receptive fields are more computationally expensive, so smaller fields  preferred. 
% of ED-TCN is

% \fakesubsection{Causal versus Acausal}
% \TODO{integrate this into other results}

% \begin{table}
% 	\centering
% 	\begin{tabular}{|l|c|c|c|}
% 		\hline
% 		  & \textbf{Seg F1} & \textbf{Edit} &  \textbf{Acc} \\
% 		\hline		
% 		WaveNet (Causal) & 86.3/64.7 &  &  \\
% 		WaveNet (Acausal) & 86.3/64.7 &  &  \\
% 		\hline        
% 		ED-TCN (Causal)  &  49.3/29.0 & 84.6/67.9 & 84.6/71.0\\
% 		ED-TCN (Acausal)  &  49.3/29.0 & 84.6/67.9 & 84.6/71.0\\        
%         \hline
% 	\end{tabular}
%             \TODO{Results: 50 Salads, video, mid?}
% 	\label{tab:shifts}	
% 	\caption{Causal verus Acausal convolutions}
% \end{table}

\fakesubsection{Training time}
It takes much less time to train a TCN than a Bi-LSTM. 
While the exact timings vary with the number of TCN layers and filter lengths, for one split of 50 Salads -- using a Nvidia Titan X for 200 epochs -- it takes about a minute to train the ED-TCN whereas and 30 minutes to train the Bi-LSTM. This speedup comes from the fact that activations within each TCN layer are all independent, and thus they can be performed in batch on a GPU. Activations in intermediate RNN layers depend on previous activations within that layer, so operations must be applied sequentially.

\section{Conclusion}
% \label{sec:discussion}

% \begin{figure*}
% 	\center
% 		\begin{tabular}{ c  c  c }
% \hspace{.1\hsize}	\textbf{50 Salads} \hspace{.1\hsize} & \hspace{.1\hsize} \textbf{MERL Shopping} \hspace{.10\hsize} & \hspace{.1\hsize} \textbf{GTEA} \hspace{.12\hsize}
% 	\end{tabular}    
%    	\includegraphics[width=\hsize]{imgs/datasets}
% %     \includegraphics[width=.45\hsize]{imgs/50Salads_timelines_models}
% % 	\includegraphics[width=.33\hsize]{imgs/50salads_timelines}
% 	\includegraphics[width=\hsize]{imgs/results}   
% 	\linebreak

% 	\caption{(top) Example images from each dataset. (bottom) Action predictions for one sequence using the mid-level action set of 50 Salads (left) and on MERL Shopping (right). These timelines are ``typical." Performance is near the average performance across each dataset.}
% 	\label{fig:timelines}
% \end{figure*}

% We performed experiments using the Time-Delay Neural Network by Waibel \etal~\cite{Waibel_1990}. We replicated it using dilated convolutions and sigmoid activation functions. 

% \fakesubsection{Conclusion}
We introduced Temporal Convolutional Networks, which use a hierarchy of convolutions to capture long-range temporal patterns. We showed on synthetic data that TCNs are capable of capturing complex patterns such as compositions, action durations, and are robust to time-delays. Our models outperformed strong baselines, including Bidirectional LSTM, and achieve state of the art performance on challenging datasets. 
We believe TCNs are a formidable alternative to RNNs and are worth further exploration. 

\fakesubsection{Acknowledgments}
Thanks to Bharat Singh and the group at MERL for discussions on their dataset and for letting us use the spatiotemporal features as input into our model. We also thank Alexander Richard for his 50 Salads predictions.
% Thanks to Hilde Kuehne and Bharat Singh for discussions on the Breakfast and MERL Shopping datasets. We greatly appreciate Kuehne and MERL for releasing the spatiotemporal features we used as input into our model.

% \TODO{Funding}

%-------------------------------------------------------------
%-------------------------------------------------------------
%In our first experiment, shown in Table~\ref{table:results1}, we compare our results with other work and perform an ablative analysis to show the performance using the spatial, the spatiotemporal, and the segmental spatiotemporal models. For clarity methods using sensor data are shown separately. These 50 Salads results are on the ``eval" granularity.
%
%In addition to evaluating our model on video data, we compare performance of the temporal component (with and without segmental inference) using the sensor data from each dataset. We also evaluate these results using LSTM, a popular Recurrent Neural Network. In these results we use 64 latent nodes.

% \input{sections/7_conclusion.tex}

% \newpage
% \input{sections/appendix.tex}

% This command serves to balance the column lengths on the last page of the document manually. 
% This command does not take effect until the next page so it should come on the page before the last.
%\addtolength{\textheight}{-12cm}   

%\section*{ACKNOWLEDGMENTS}
%Funding and personal acknowledgments will be added upon acceptance

\newpage
{\small
\bibliographystyle{ieee}
%\bibliography{/Users/colin/Documents/library,../bib/activity_recognition,../bib/ColinLea}
\bibliography{bib/ColinLea.bib,bib/activity_recognition.bib,bib/new.bib}

\begin{thebibliography}{10}\itemsep=-1pt

\bibitem{tensorflow}
M.~Abadi, A.~Agarwal, P.~Barham, E.~Brevdo, Z.~Chen, C.~Citro, G.~S. Corrado,
  A.~Davis, J.~Dean, M.~Devin, S.~Ghemawat, I.~Goodfellow, A.~Harp, G.~Irving,
  M.~Isard, Y.~Jia, R.~Jozefowicz, L.~Kaiser, M.~Kudlur, J.~Levenberg,
  D.~Man\'{e}, R.~Monga, S.~Moore, D.~Murray, C.~Olah, M.~Schuster, J.~Shlens,
  B.~Steiner, I.~Sutskever, K.~Talwar, P.~Tucker, V.~Vanhoucke, V.~Vasudevan,
  F.~Vi\'{e}gas, O.~Vinyals, P.~Warden, M.~Wattenberg, M.~Wicke, Y.~Yu, and
  X.~Zheng.
\newblock {TensorFlow}: Large-scale machine learning on heterogeneous systems,
  2015.
\newblock Software available from tensorflow.org.

\bibitem{badrinarayanan_arxiv_2015}
V.~Badrinarayanan, A.~Handa, and R.~Cipolla.
\newblock Segnet: A deep convolutional encoder-decoder architecture for robust
  semantic pixel-wise labelling.
\newblock {\em arXiv preprint arXiv:1505.07293}, 2015.

\bibitem{cheron_iccv_2015}
G.~Cheron, I.~Laptev, and C.~Schmid.
\newblock P-cnn: Pose-based cnn features for action recognition.
\newblock 2015.

\bibitem{keras}
F.~Chollet.
\newblock Keras.
\newblock \url{https://github.com/fchollet/keras}, 2015.

\bibitem{bi_lstm}
W.~Duch, J.~Kacprzyk, E.~Oja, and S.~Zadrozny, editors.
\newblock {\em Artificial Neural Networks: Formal Models and Their Applications
  - {ICANN} 2005, 15th International Conference, Warsaw, Poland, September
  11-15, 2005, Proceedings, Part {II}}, volume 3697 of {\em Lecture Notes in
  Computer Science}. Springer, 2005.

\bibitem{fathi_iccv_2011}
A.~Fathi, A.~Farhadi, and J.~M. Rehg.
\newblock Understanding egocentric activities.
\newblock 2011.

\bibitem{fathi_cvpr_2013}
A.~Fathi and J.~M. Rehg.
\newblock Modeling actions through state changes.
\newblock 2013.

\bibitem{fathi_cvpr_2011}
A.~Fathi, R.~Xiaofeng, and J.~M. Rehg.
\newblock Learning to recognize objects in egocentric activities.
\newblock 2011.

\bibitem{huang_eccv_2016}
D.-A. Huang, L.~Fei-Fei, and J.~C. Niebles.
\newblock {\em Connectionist Temporal Modeling for Weakly Supervised Action
  Labeling}, pages 137--153.
\newblock Springer International Publishing, Cham, 2016.

\bibitem{jain_cvpr_2015}
M.~Jain, J.~C. van Gemert, and C.~G.~M. Snoek.
\newblock What do 15,000 object categories tell us about classifying and
  localizing actions?
\newblock 2015.

\bibitem{karpathy_cvpr_2014}
A.~Karpathy, G.~Toderici, S.~Shetty, T.~Leung, R.~Sukthankar, and L.~Fei-Fei.
\newblock Large-scale video classification with convolutional neural networks.
\newblock 2014.

\bibitem{ADAM}
D.~P. Kingma and J.~Ba.
\newblock Adam: {A} method for stochastic optimization.
\newblock 2014.

\bibitem{kuehne_cvpr_2014}
H.~Kuehne, A.~Arslan, and T.~Serre.
\newblock The language of actions: Recovering the syntax and semantics of
  goal-directed human activities.
\newblock In {\em The IEEE Conference on Computer Vision and Pattern
  Recognition (CVPR)}, June 2014.

\bibitem{kuehne_wacv_2016}
H.~Kuehne, J.~Gall, and T.~Serre.
\newblock An end-to-end generative framework for video segmentation and
  recognition.
\newblock Lake Placid, Mar 2016.

\bibitem{lea_eccv_2016}
C.~Lea, A.~Reiter, R.~Vidal, and G.~D. Hager.
\newblock Segmental spatio-temporal {CNNs} for fine-grained action
  segmentation.
\newblock 2016.

\bibitem{lea_icra_2016}
C.~Lea, R.~Vidal, and G.~D. Hager.
\newblock Learning convolutional action primitives for fine-grained action
  recognition.
\newblock 2016.

\bibitem{TDD}
Y.~Q. Limin~Wang and X.~Tang.
\newblock Action recognition with trajectory-pooled deep-convolutional
  descriptors.
\newblock 2015.

\bibitem{ng_cvpr_2015}
J.~Y. Ng, M.~J. Hausknecht, S.~Vijayanarasimhan, O.~Vinyals, R.~Monga, and
  G.~Toderici.
\newblock Beyond short snippets: Deep networks for video classification.
\newblock 2015.

\bibitem{ni_cvpr_2014}
B.~Ni, V.~R. Paramathayalan, and P.~Moulin.
\newblock Multiple granularity analysis for fine-grained action detection.
\newblock 2014.

\bibitem{ni_cvpr_2016}
B.~Ni, X.~Yang, and S.~Gao.
\newblock Progressively parsing interactional objects for fine grained action
  detection.
\newblock In {\em The IEEE Conference on Computer Vision and Pattern
  Recognition (CVPR)}, June 2016.

\bibitem{pascanu_icml_2013}
R.~Pascanu, T.~Mikolov, and Y.~Bengio.
\newblock On the difficulty of training recurrent neural networks.
\newblock 2013.

\bibitem{peng_thumos_2015}
X.~Peng and C.~Schmid.
\newblock Encoding feature maps of cnns for action recognition.
\newblock In {\em CVPR, THUMOS Challenge 2015 Workshop}, 2015.

\bibitem{pirsiavash_cvpr_2014}
H.~Pirsiavash and D.~Ramanan.
\newblock Parsing videos of actions with segmental grammars.
\newblock 2014.

\bibitem{richard_cvpr_2016}
A.~Richard and J.~Gall.
\newblock Temporal action detection using a statistical language model.
\newblock 2016.

\bibitem{rohrbach_ijcv_2015}
M.~Rohrbach, A.~Rohrbach, M.~Regneri, S.~Amin, M.~Andriluka, M.~Pinkal, and
  B.~Schiele.
\newblock Recognizing fine-grained and composite activities using hand-centric
  features and script data.
\newblock 2015.

\bibitem{simonyan_nips_2014}
K.~Simonyan and A.~Zisserman.
\newblock Two-stream convolutional networks for action recognition in videos.
\newblock 2014.

\bibitem{singh_cvpr_2016_merl}
B.~Singh, T.~K. Marks, M.~Jones, O.~Tuzel, and M.~Shao.
\newblock A multi-stream bi-directional recurrent neural network for
  fine-grained action detection.
\newblock 2016.

\bibitem{singh_cvpr_2016_ego}
S.~Singh, C.~Arora, and C.~V. Jawahar.
\newblock First person action recognition using deep learned descriptors.
\newblock June 2016.

\bibitem{stein_ubicomp_2013}
S.~Stein and S.~J. McKenna.
\newblock Combining embedded accelerometers with computer vision for
  recognizing food preparation activities.
\newblock 2013.

\bibitem{sun_iccv_2015}
L.~Sun, K.~Jia, D.-Y. Yeung, and B.~Shi.
\newblock Human action recognition using factorized spatio-temporal
  convolutional networks.
\newblock 2015.

\bibitem{tompson_cvpr_2015}
J.~Tompson, R.~Goroshin, A.~Jain, Y.~LeCun, and C.~Bregler.
\newblock Efficient object localization using convolutional networks.
\newblock In {\em The IEEE Conference on Computer Vision and Pattern
  Recognition (CVPR)}, June 2015.

\bibitem{tran_iccv_2015}
D.~Tran, L.~Bourdev, R.~Fergus, L.~Torresani, and M.~Paluri.
\newblock Learning spatiotemporal features with 3d convolutional networks.
\newblock 2015.

\bibitem{wavenet}
A.~van~den Oord, S.~Dieleman, H.~Zen, K.~Simonyan, O.~Vinyals, A.~Graves,
  N.~Kalchbrenner, A.~W. Senior, and K.~Kavukcuoglu.
\newblock Wavenet: {A} generative model for raw audio.
\newblock {\em CoRR}, abs/1609.03499, 2016.

\bibitem{GatedPixelCNN}
A.~van~den Oord, N.~Kalchbrenner, O.~Vinyals, L.~Espeholt, A.~Graves, and
  K.~Kavukcuoglu.
\newblock Conditional image generation with pixelcnn decoders.
\newblock {\em CoRR}, abs/1606.05328, 2016.

\bibitem{Waibel_1990}
A.~Waibel, T.~Hanazawa, G.~Hinton, K.~Shikano, and K.~J. Lang.
\newblock Readings in speech recognition.
\newblock chapter Phoneme Recognition Using Time-delay Neural Networks, pages
  393--404. Morgan Kaufmann Publishers Inc., San Francisco, CA, USA, 1990.

\bibitem{wang_iccv_2013}
H.~Wang and C.~Schmid.
\newblock Action recognition with improved trajectories.
\newblock 2013.

\bibitem{yu_iclr_2016}
F.~Yu and V.~Koltun.
\newblock Multi-scale context aggregation by dilated convolutions.
\newblock In {\em ICLR}, 2016.

\bibitem{zhou_cvpr_2015}
Y.~Zhou, B.~Ni, R.~Hong, M.~Wang, and Q.~Tian.
\newblock Interaction part mining: A mid-level approach for fine-grained action
  recognition.
\newblock In {\em The IEEE Conference on Computer Vision and Pattern
  Recognition (CVPR)}, June 2015.

\end{thebibliography}
}

\end{document}